\documentclass[conference]{IEEEtran}

\usepackage[T1]{fontenc}
\usepackage[utf8]{inputenc}
\usepackage{newpxtext}
\usepackage{newpxmath}
\usepackage{microtype}
\usepackage{graphicx}
\usepackage{booktabs}
\usepackage{amsmath}
\usepackage{url}
\usepackage[hidelinks]{hyperref}
\usepackage{xcolor}

\newcommand{\hflink}[1]{\url{https://huggingface.co/datasets/#1}}

\begin{document}

\title{Open Annotations and Synthetic Data for Field Localisation in
Indian Bank Cheques}

\author{\IEEEauthorblockN{Jaganadh Gopinadhan}
\IEEEauthorblockA{\textit{Senior Member, IEEE}\\
jgopinadhan@acm.org}}

\maketitle

\begin{abstract}
Automated cheque processing requires localising key fields (date, legal
amount, IFSC code, account number, signature, and payee name) before any
recognition step. The IDRBT Cheque Image Dataset is, to our knowledge, the
only public collection of Indian bank cheques, but it ships without field
annotations and with no stated licence, so its redistribution terms are
unclear. We address both limitations. First, we release
six-field bounding-box annotations for all 112 cheques in the dataset,
distributed annotations-only and keyed to the original filenames so that the
IDRBT redistribution terms are respected. Second, we release 295 fully
redistributable synthetic cheque images produced by a cut-paste pipeline that
composites annotated field regions from real cheques onto content-erased,
bank-specific canvas templates; because patches are pasted at their source
coordinates, annotations carry forward unchanged. Third, we provide a
ResNet-50 direct-regression baseline that predicts all six fields in a single
forward pass, and use it for a controlled test of the synthetic data. The
test is sobering: because cheque layouts are rigid, a no-learning baseline
that simply predicts each field's mean training box already reaches 0.691
mean IoU and 80\% accuracy at IoU\,$\geq$\,0.5, and once seed variance and
training compute are accounted for, the cut-paste synthetic data yields no
measurable improvement over real data alone (an equal-compute real-only
model matches or beats the synthetic-augmented model on every aggregate
metric). We report this negative result in full, since it cautions against
assuming appearance-only augmentation helps fixed-layout documents and
points instead to layout-varying synthesis. The annotations and synthetic
images are released as reusable resources on the Hugging Face Hub under
permissive licences.
\end{abstract}

\section{Introduction}
\label{sec:intro}

Bank cheques remain a significant payment instrument in India, and automated
cheque truncation systems must locate the handwritten and printed fields of a
cheque before optical character recognition or signature verification can be
applied. Although document-understanding research has produced rich annotated
benchmarks for forms and receipts, cheques are essentially absent from public
datasets: banking documents are privacy-sensitive and rarely releasable.

The IDRBT Cheque Image Dataset~\cite{idrbt2020, dansena2017} is the notable
exception: a public set of 112 scanned Indian bank cheques from four banks,
originally assembled for pen-ink differentiation research~\cite{dansena2017}.
Two gaps, however, have limited its use for machine learning. The dataset
provides no field-level annotations, and it is distributed with no stated
licence, so its redistribution terms are unclear (a third party has
nonetheless mirrored the images on
Kaggle\footnote{\url{https://www.kaggle.com/datasets/jdranpariya/cheque-data}},
which makes them easy to obtain but does not clarify the licensing). This
paper fills both gaps with three resources:

\begin{itemize}
  \item \textbf{Field annotations.} Six bounding boxes (date, amount, IFSC,
  account number, signature, payee name) for each of the 112 cheques,
  released as an annotations-only dataset keyed to the original IDRBT
  filenames (\hflink{jaganadhg/cheque-field-annotations}).
  \item \textbf{Redistributable synthetic images.} A cut-paste generation
  pipeline, adapted from Dwibedi et~al.~\cite{dwibedi2017cut} to
  semi-structured documents, yielding 295 synthetic cheques with
  carried-forward annotations (\hflink{jaganadhg/cheque-synthetic-images}).
  No synthetic image reproduces a complete original document.
  \item \textbf{A baseline and a controlled negative result.} A ResNet-50
  direct-regression model that predicts all six fields in one forward
  pass\footnote{Code: \url{https://github.com/jaganadhg/finimgproc}}
  (\url{https://huggingface.co/jaganadhg/cheque-field-regressor}), used to
  test whether the synthetic data helps. Under a controlled comparison
  (three seeds, plus a compute-matched real-only control) it does not, and
  we report and analyse this negative result rather than the most
  favourable single run.
\end{itemize}

\section{Related Work}
\label{sec:related}

\paragraph{Cheque processing} Classical cheque automation systems rely on
template matching or bank-specific zone rules to extract fields, which are
brittle to layout, print, and scan variation. Deep-learning pipelines have
since been applied to end-to-end cheque verification: Agrawal
et~al.~\cite{agrawal2021cheque} extract cheque components (amounts, account
number, signature) with image processing and CNNs, Chaitanyaswami
et~al.~\cite{chaitanyaswami2025cheque} address overlapping and faded
handwriting on cheques with a multi-stage recognition framework, and Singh
et~al.~\cite{singh2024cheque} and Pavan Kumar et~al.~\cite{pavankumar2026}
build verification pipelines on the IDRBT cheque dataset itself, combining
CNN-based handwriting recognition with OCR and SIFT/SVM signature matching;
industry pipelines follow the same template~\cite{ignitarium_chequeblog}.
All such systems presuppose a field-localisation stage, yet none releases
its field annotations; this is precisely the gap the present resources
address.

\paragraph{Field localisation on IDRBT} Closest to our baseline task, a few
works localise cheque fields on the IDRBT images directly. Abdo
et~al.~\cite{abdo2023} train a Faster R-CNN detector and report 97.4\%
field-detection accuracy, and a community project~\cite{pranav_chequedetection}
re-annotates the dataset (with the SuperAnnotate tool) and trains an SSD
MobileNet detector. These report strong single-run numbers but, like the
verification pipelines above, neither releases its field annotations nor
benchmarks against a trivial layout prior or reports seed variance. Our
controlled study (Section~\ref{sec:results}) supplies exactly that context:
on this dataset a no-learning predictor of each field's mean box already
scores highly, so headline accuracies on IDRBT should be read against that
prior. We stress that building any of these downstream systems (verification,
recognition, OCR, fraud detection) is \emph{not} our objective; we release
reusable annotations and synthetic data and characterise the localisation
task honestly.

\paragraph{Document layout datasets} FUNSD~\cite{jaume2019funsd} (forms),
SROIE~\cite{huang2019sroie} (scanned receipts), and CORD~\cite{park2019cord}
(receipts) anchor research on field detection in semi-structured documents.
None of these covers bank cheques, whose mixture of fixed printed structure
and free handwriting (amounts, signatures) is distinctive.

\paragraph{Cheque datasets} The closest public resource is
BCSD~\cite{khan2021bcsd}, a segmentation dataset for \emph{signatures} on
bank cheques, with pixel-level and patch-level masks over cheque images
gathered from mixed public and scanned sources. BCSD targets a single field
and the segmentation task; our annotations instead cover six fields with
bounding boxes on the established IDRBT benchmark, and our synthetic dataset
additionally provides redistributable full-cheque training images. The two
resources are complementary; BCSD's signature masks could, for instance,
refine the coarse signature boxes predicted by our baseline.

\paragraph{Cut-paste synthesis} Dwibedi et~al.~\cite{dwibedi2017cut} showed
that pasting object instances onto new backgrounds, even without
blending, produces effective training data for instance detection. We adapt
the idea to fixed-layout documents: instead of pasting at random positions,
field crops are pasted at their source coordinates onto a content-erased
canvas of the same bank format, preserving the layout prior that the
localisation model must learn.

\section{The Annotated Dataset}
\label{sec:dataset}

\subsection{Source images}

The IDRBT Cheque Image Dataset~\cite{idrbt2020} contains 112 cheque scans
from four Indian banks: Axis (87), Canara (10), ICICI (8), and Syndicate (7).
Images are RGB TIFFs of approximately $2365 \times 1087$ pixels at roughly
300\,DPI (A5 landscape). The bank distribution is heavily skewed: Axis alone
accounts for 78\% of the images. The dataset was created at IDRBT
specifically for research, originally to study pen-ink
differentiation~\cite{dansena2017}: nine volunteers wrote on cheque leaves
from the four banks using fourteen pens (seven blue, seven black) to
diversify handwriting and ink, each cheque being written by two volunteers
with two different pens, and the leaves were scanned on an ordinary flatbed
scanner~\cite{idrbt2020}. It is openly downloadable but carries no stated
licence.

\subsection{Annotation protocol}

Each cheque was annotated with six axis-aligned bounding boxes: \emph{date},
\emph{amount} (courtesy amount box), \emph{ifsc} (printed IFSC code),
\emph{acno} (account number), \emph{sign} (signature region), and \emph{name}
(payee line). All 112 cheques were annotated manually by the author using
LabelImg~\cite{labelimg}, circa 2020, with the boxes stored per-image in
Pascal VOC XML and consolidated for release. As a
quality-control step, the author performed two full rounds of visual
verification, overlaying every bounding box on its source image and correcting
discrepancies.

\begin{figure}[t]
  \centering
  \includegraphics[width=\linewidth]{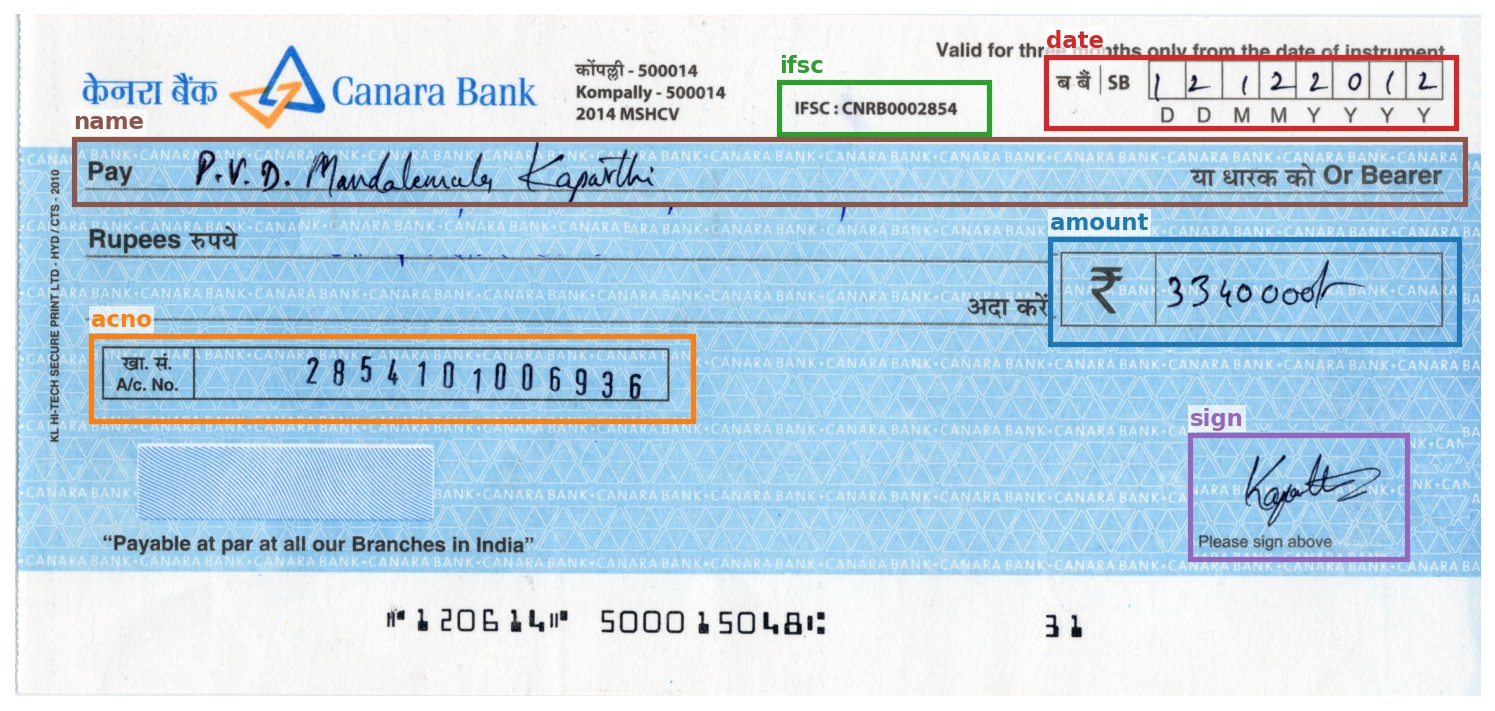}
  \caption{The six annotated fields, shown on a \emph{synthetic} cheque from
  our generated dataset (the source IDRBT images carry no explicit licence,
  so we show a synthetic one).}
  \label{fig:layout}
\end{figure}

Figure~\ref{fig:layout} shows the six fields on a synthetic cheque. Every
cheque contains exactly one instance of each field, a property the baseline
model exploits (Section~\ref{sec:baseline}).

\subsection{Field layout statistics}

Table~\ref{tab:bbox_stats} summarises the spatial distribution of the
annotated boxes in normalised coordinates. Most fields have low positional
variance ($\sigma < 0.05$), reflecting the semi-structured nature of cheques;
this motivates both the same-coordinate paste of
Section~\ref{sec:synthetic} and the layout-aware regression head of
Section~\ref{sec:baseline}. The \emph{ifsc} field is the smallest (4.7\% of
image height); it is an important field on Indian cheques, as the IFSC code
identifies the issuing bank branch, yet in the real images it sits within a
text-dense region (printed branch address, MICR line, and form labels), which
makes it hard to delimit. The \emph{sign} field has the largest size
variance, since signature extent depends on the account holder.

\begin{table}[t]
  \caption{Per-field bounding-box statistics over the 112 annotated cheques
  (normalised coordinates, mean values; $\sigma$ = std.\ dev.).}
  \label{tab:bbox_stats}
  \centering
  \small
  \begin{tabular}{lcccccc}
    \toprule
    Field & Ctr-$x$ & Ctr-$y$ & W & H & W$\sigma$ & H$\sigma$ \\
    \midrule
    date   & 0.854 & 0.125 & 0.264 & 0.136 & 0.015 & 0.014 \\
    amount & 0.846 & 0.424 & 0.269 & 0.135 & 0.008 & 0.014 \\
    ifsc   & 0.224 & 0.157 & 0.125 & 0.047 & 0.009 & 0.009 \\
    acno   & 0.211 & 0.529 & 0.317 & 0.107 & 0.053 & 0.017 \\
    sign   & 0.904 & 0.721 & 0.140 & 0.221 & 0.024 & 0.039 \\
    name   & 0.510 & 0.251 & 0.960 & 0.109 & 0.009 & 0.013 \\
    \bottomrule
  \end{tabular}
\end{table}

\subsection{Release format and data quality}

The annotations are released as a Hugging Face dataset containing one record
per cheque (filename, bank, image size, and the six boxes) \emph{without}
the images; since the source images carry no explicit licence, we release
only our own annotations and let users obtain the TIFFs from IDRBT and join
on filename. The annotation records themselves are released under the dataset card's
terms, which defer to the IDRBT licence for the underlying images; the
synthetic dataset (Section~\ref{sec:synthetic}) is the Apache-2.0 release.
All 112 records are complete, with a 90/11/11
train/validation/test split, which we designate as the canonical benchmark
split for future work. An earlier HDF5 consolidation of the annotations
corrupted seven rows to NaN; the regression baseline in
Section~\ref{sec:baseline} predates the fix and was trained on the 105
uncorrupted records with an 85/10/10 split. Its numbers are therefore
indicative rather than canonical, and re-establishing the baseline on the
released split is immediate future work.

\subsection{Independent annotation agreement}
\label{sec:agreement}

To gauge the quality and the inherent ambiguity of the annotations, we
compare them against an independent community annotation of the same IDRBT
images~\cite{pranav_chequedetection}, produced with a different tool
(SuperAnnotate). The two label sets share four fields (\emph{date},
\emph{amount}, \emph{acno}, \emph{name}); the community set instead includes
a cheque-number and an issuing-bank box, and notably annotates neither
\emph{ifsc} nor \emph{sign}. Our release thus adds exactly the two fields
most tied to downstream cheque processing: \emph{ifsc} localises the printed
IFSC code that identifies the issuing bank branch, and \emph{sign} localises
the signature region that is central to authenticity verification.

Table~\ref{tab:agreement} reports the per-field IoU between the two
annotators over all 112 images. Agreement is highest on the crisp printed
\emph{date} and \emph{amount} boxes (0.71) and lowest on the payee
\emph{name} line (0.54), whose horizontal extent is inherently ambiguous;
the overall mean is 0.65, and the two annotators place boxes within IoU
$\geq 0.5$ of each other 86.8\% of the time. That an independent annotator
using a different tool produced compatible boxes corroborates the annotation
quality. It also establishes a practical \emph{ceiling}: ``ground truth''
for cheque fields carries roughly 0.35 IoU of annotator-to-annotator slack,
so absolute IoU values near 0.65--0.70 are close to the achievable maximum,
and IoU differences smaller than this scale should not be over-interpreted.
We return to this point in Section~\ref{sec:results}. Two caveats: the gap is
partly definitional rather than error (e.g.\ the community \emph{acno} boxes
are systematically wider, including the printed ``A/c No.'' label), so 0.65
is a conservative ceiling; and this is pairwise agreement between two
annotators, not a full multi-annotator study.

\begin{table}[t]
  \caption{Per-field agreement (IoU) between our annotations and an
  independent community annotation~\cite{pranav_chequedetection} of the same
  112 cheques, over the four shared fields. \emph{ifsc} and \emph{sign} are
  unique to our release and have no independent counterpart.}
  \label{tab:agreement}
  \centering
  \small
  \begin{tabular}{lcccc}
    \toprule
    Field & $n$ & mean IoU & median & \%\,IoU$\geq$0.5 \\
    \midrule
    date   & 112 & 0.712 & 0.734 & 94.6\% \\
    amount & 112 & 0.712 & 0.725 & 93.8\% \\
    acno   & 111 & 0.633 & 0.629 & 87.4\% \\
    name   & 112 & 0.543 & 0.553 & 71.4\% \\
    \midrule
    All    & 447 & 0.650 & 0.642 & 86.8\% \\
    \bottomrule
  \end{tabular}
\end{table}

\section{Synthetic Data Generation}
\label{sec:synthetic}

\subsection{Motivation}

Three properties of the real data motivate synthesis: the dataset is small
(112 images), severely imbalanced across banks (78\% Axis), and carries no
explicit licence. A synthetic dataset addresses all three: it can be openly
licensed, and its annotations are correct by construction. The pipeline below is, above
all, a way to turn a handful of real cheques into an arbitrarily large
corpus of redistributable, fully-annotated images; this scaling property is
the contribution, independent of whether the resulting images improve any
particular model (a question we examine, and do not settle in the
synthetic data's favour, in Section~\ref{sec:results}).

\begin{figure}[t]
  \centering
  \includegraphics[width=\linewidth]{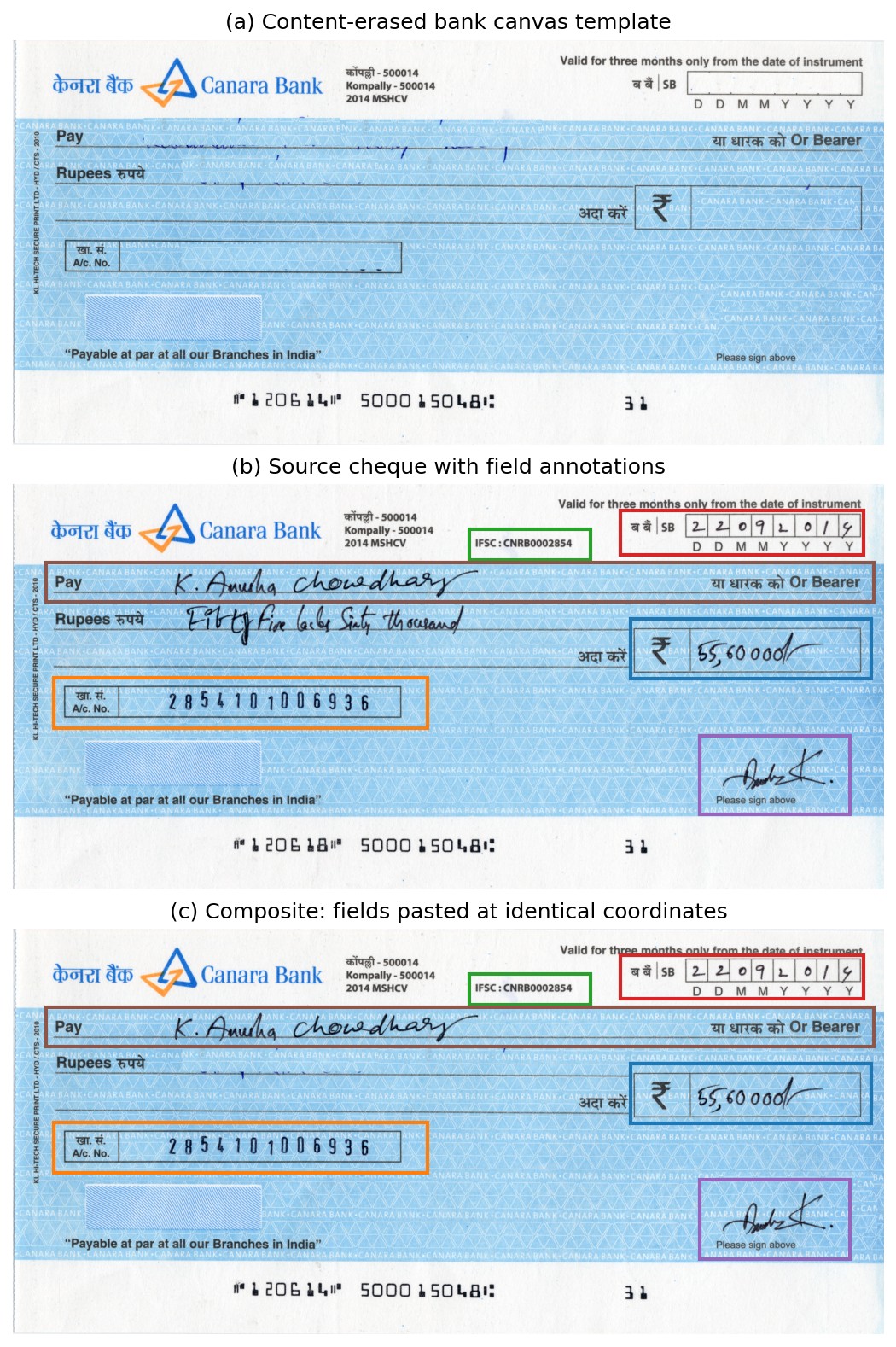}
  \caption{Cut-paste generation. (a)~Content-erased canvas template for the
  bank. (b)~A source cheque with its six annotated fields. (c)~The composite:
  each field crop is pasted onto the canvas at its source coordinates, so the
  annotations carry forward unchanged.}
  \label{fig:pipeline}
\end{figure}

\subsection{Method}

For each of the four banks, one \emph{canvas template} was prepared by
manually erasing all handwritten and printed field content from a real cheque
of that bank, leaving only static structure (borders, logos, form labels). To
generate a synthetic image (Fig.~\ref{fig:pipeline}): pick a bank; pick a
random source cheque of that bank; crop its six annotated field regions; and
paste each crop onto a fresh copy of the bank's canvas at the \emph{same}
pixel coordinates. The source annotations are then valid for the composite
without modification.

Unlike Dwibedi et~al.~\cite{dwibedi2017cut}, we do not randomise paste
positions: field placement on a cheque is a strong prior that the downstream
localisation model must learn, and random repositioning would destroy it. The
synthesis therefore adds \emph{appearance} diversity (handwriting, ink,
content) while preserving \emph{layout} statistics: the synthetic
bounding-box statistics closely match Table~\ref{tab:bbox_stats}
(Appendix~\ref{app:fidelity}).

\subsection{Generated dataset}

With seed 42, Axis was generated at a $1{\times}$ ratio and the three small
banks at $10{\times}$ to counter the bank imbalance: 337 images were
attempted and 295 generated (12\% loss from source TIFFs referenced in the
metadata but absent from the release). Table~\ref{tab:synth_counts} shows the
per-bank counts and splits.

\begin{table}[t]
  \caption{Real vs.\ synthetic image counts and synthetic splits.}
  \label{tab:synth_counts}
  \centering
  \small
  \begin{tabular}{lccccc}
    \toprule
    Bank & Real & Synthetic & Train & Val & Test \\
    \midrule
    Axis      & 87 & 79 & 65 & 6 & 8 \\
    Canara    & 10 & 91 & 78 & 5 & 8 \\
    ICICI     &  8 & 63 & 51 & 6 & 6 \\
    Syndicate &  7 & 62 & 41 & 13 & 8 \\
    \midrule
    Total     & 112 & 295 & 235 & 30 & 30 \\
    \bottomrule
  \end{tabular}
\end{table}

\subsection{Redistribution}

Each composite combines field crops from one source cheque with the erased
canvas of its bank; no synthetic image reproduces a complete original
document, and the static background is a manually content-erased derivative.
The synthetic dataset is therefore released in full
(images and annotations) under Apache-2.0, making it, to our knowledge, the
first freely redistributable cheque image dataset with full-field
annotations (BCSD~\cite{khan2021bcsd} releases signature masks only).

A note on personal data: the pasted crops include handwritten signatures,
account numbers, and payee names, which would be sensitive if they belonged
to real customers. They do not: as described in
Section~\ref{sec:dataset}, the IDRBT cheques were written by nine
volunteers expressly to create a research dataset~\cite{idrbt2020}, so the
fields are not records of customer transactions and do not correspond to
live accounts or real account holders. We nonetheless treat the signature
crops conservatively and note that the composites reproduce the volunteers'
handwriting verbatim.

\subsection{Limitations}

Paste positions are not jittered, so within a bank the synthetic data adds no
layout diversity; patches are pasted with hard seams (no blending); small
banks sample source cheques with replacement, so content repeats; and a
single canvas per bank cannot represent intra-bank sub-formats. The first of
these turns out to matter most: the controlled experiment of
Section~\ref{sec:results} finds no measurable localisation benefit from the
synthetic data, and we argue there that the absence of \emph{layout}
diversity, not appearance, is the likely reason.

More broadly, format independence is a hard open problem for Indian cheques:
colour schemes, typefaces, and background shading vary substantially across
issuing banks, so a single canvas template per bank captures only a fraction
of the real visual diversity. Generalising across this variety, rather than
augmenting appearance within a fixed layout, is the harder challenge our
results point to.

\section{Baseline Model}
\label{sec:baseline}

We emphasise that producing the strongest possible model is not the
objective of this work; the annotations of Section~\ref{sec:dataset} and the
synthetic-generation approach of Section~\ref{sec:synthetic} are the
contributions. The baseline exists to demonstrate that the released
annotations support training and, above all, to test the synthetic data
with a controlled experiment. We nonetheless report the full v1--v4
evolution, including missteps, since the lessons transfer to anyone training
on these resources.

\subsection{Direct regression formulation}

Every cheque contains exactly one instance of each of the six fields, so
detection machinery is unnecessary: given an image, the model directly
regresses a $6 \times 4$ tensor of normalised
$[x_{\min}, y_{\min}, x_{\max}, y_{\max}]$ boxes. We prototyped
DETR~\cite{carion2020detr} and Faster R-CNN~\cite{ren2015fasterrcnn}
alternatives; with $\sim$100 training images, Hungarian-matching instability
(DETR) and anchor tuning on an over-parameterised head (Faster R-CNN)
prevented reliable convergence, while regression with a SmoothL1 loss trains
stably and needs no confidence thresholding or NMS at inference.

\subsection{Architecture}

The backbone is an ImageNet-pretrained ResNet-50~\cite{he2016resnet} at
$1024 \times 512$ input resolution. Features from \texttt{layer3} and
\texttt{layer4} are projected to 128 channels by $1{\times}1$ convolutions
and pooled to a $4 \times 2$ spatial grid each; the concatenated 2048-d
vector feeds a two-layer head (FC 512, dropout 0.3, FC 24, sigmoid). The
critical choice is the $4 \times 2$ \emph{spatial} pool in place of global
average pooling: GAP discards the positional information that localisation
needs, and restoring a coarse layout grid is the largest single
\emph{architectural} improvement in our ablation (+0.15 mIoU,
Table~\ref{tab:ablation}), though as Section~\ref{sec:results} shows, even
this does not lift the model past the no-learning layout prior.

Training uses a field-weighted SmoothL1 loss (weights 2.0 for \emph{ifsc},
1.5 for \emph{sign}, 1.0 otherwise), AdamW~\cite{loshchilov2019adamw}, and a
two-phase schedule: 15 epochs with the backbone frozen at learning rate
$10^{-4}$, then end-to-end at $10^{-5}$ with cosine annealing, 150 epochs
total. Online augmentations (horizontal flip, $\pm 5^{\circ}$ affine, colour
jitter, Gaussian blur) are applied jointly to images and boxes with
torchvision \texttt{transforms.v2}. Full reproducibility details (commands,
seeds, hardware, checkpoint provenance) are given in
Appendix~\ref{app:repro}.

\subsection{Training with synthetic data}

The v4 model adds the synthetic training split to the 85 real training
images. To prevent leakage, we recover the source cheque of every synthetic
image by matching its carried-forward box coordinates against the real
annotations, and exclude the 41 synthetic images whose source cheque lies in
the real validation or test split, leaving 194 synthetic images (279 training
images in total). The canvas templates themselves derive from one cheque per bank, but all
field content is erased from them; only printed structure is shared, which
is common to every cheque of that bank, so template provenance cannot leak
validation or test content. Validation and test sets remain real images only,
identical across all variants.

Because the test set is small, a single training run is not a sound basis for
a comparison. We therefore evaluate the synthetic data with a controlled
protocol: the synthetic-augmented model (v4) is trained under three random
seeds that vary initialisation and augmentation while holding the data split
fixed, and we add a \emph{compute-matched} real-only control, v3 trained for
495 epochs so that it sees the same number of gradient steps as v4 does in
150 epochs over the $3.3\times$ larger combined set. This isolates the effect
of the synthetic \emph{data} from the confounds of seed luck and training
length.

\subsection{Results}
\label{sec:results}

\begin{table}[t]
  \caption{Single-seed architecture ablation on the held-out real test set
  (10 images), against a no-learning \emph{static} baseline that predicts
  the per-field mean training box. IoU is the per-field mean; Acc is
  accuracy at IoU $\geq 0.5$. v1: GAP head, $800{\times}400$. v2: +
  resolution $1024{\times}512$, affine aug., aggressive field weights. v3: +
  $4{\times}2$ spatial pooling, moderate weights. v4: v3 trained with real +
  synthetic data. These are single runs; the v4 column ($\dag$) is the best
  of three seeds, and Table~\ref{tab:controlled} reports the proper
  seed-averaged and compute-matched comparison.}
  \label{tab:ablation}
  \centering
  \scriptsize
  \setlength{\tabcolsep}{2.6pt}
  \begin{tabular}{lcccccccccc}
    \toprule
    & \multicolumn{2}{c}{static} & \multicolumn{2}{c}{v1} & \multicolumn{2}{c}{v2}
    & \multicolumn{2}{c}{v3} & \multicolumn{2}{c}{v4 (+synth)$^{\dag}$} \\
    Field & IoU & Acc & IoU & Acc & IoU & Acc & IoU & Acc & IoU & Acc \\
    \midrule
    date   & 0.827 & 100\% & 0.455 & 50\% & 0.459 & 50\% & 0.528 & 50\% & 0.682 & 100\% \\
    amount & 0.859 & 100\% & 0.537 & 70\% & 0.408 & 20\% & 0.572 & 80\% & 0.724 & 100\% \\
    ifsc   & 0.296 &   0\% & 0.039 &  0\% & 0.334 & 20\% & 0.506 & 60\% & 0.678 & 90\% \\
    acno   & 0.743 & 100\% & 0.482 & 70\% & 0.433 & 20\% & 0.579 & 90\% & 0.715 & 100\% \\
    sign   & 0.614 &  80\% & 0.154 &  0\% & 0.290 &  0\% & 0.437 & 30\% & 0.614 & 100\% \\
    name   & 0.807 & 100\% & 0.632 & 80\% & 0.446 & 30\% & 0.658 & 90\% & 0.808 & 100\% \\
    \midrule
    Mean   & 0.691 &  80\% & 0.383 & 45\% & 0.395 & 23\% & 0.547 & 67\% & 0.704 & 98\% \\
    \bottomrule
  \end{tabular}
\end{table}

\begin{table}[t]
  \caption{Controlled comparison on the real test set. The static prior, v3,
  and v3-long are single runs; v4 is reported as mean $\pm$ standard
  deviation over three seeds. v3-long is v3 trained for 495 epochs, matching
  v4's total gradient steps. The compute-matched real-only model (v3-long)
  matches or beats the synthetic-augmented model (v4) on every metric, and
  no learned model beats the static prior on mIoU.}
  \label{tab:controlled}
  \centering
  \small
  \setlength{\tabcolsep}{4pt}
  \begin{tabular}{llccc}
    \toprule
    Model & Training data & mIoU & mAcc@0.5 & MAE \\
    \midrule
    Static prior  & none              & 0.691 & 80.0\% & 0.0174 \\
    v3            & real, 150 ep      & 0.547 & 66.7\% & 0.0270 \\
    v3-long       & real, 495 ep      & 0.660 & 93.3\% & 0.0179 \\
    v4            & real+synth, 150 ep & $0.648$ & $89.4\%$ & $0.0185$ \\
    (3 seeds)     &                   & ${\scriptstyle \pm 0.041}$ & ${\scriptstyle \pm 7.5}$ & ${\scriptstyle \pm 0.002}$ \\
    \bottomrule
  \end{tabular}
\end{table}

\paragraph{The layout prior is a strong baseline} The static column of
Table~\ref{tab:ablation} is the essential context for everything that
follows: predicting the per-field mean training box, with no learning at
all, reaches 0.691 mIoU and 80\% accuracy. Cheque layout is so regular that
this trivial predictor is competitive, and \emph{no} learned variant in
this paper exceeds it on mean IoU. Learned models must therefore be judged
by their margin over the prior, not by absolute IoU. This is reinforced from
the annotation side: independent cross-annotator agreement is itself only
0.65 mean IoU (Section~\ref{sec:agreement}), so the IoU values here sit near
the annotation ceiling and inter-column IoU gaps below that scale fall within
labelling noise. Accuracy at IoU $\geq 0.5$, where two annotators concur
87\% of the time, is the more trustworthy metric.

\paragraph{Architecture trends} Read as single-seed exploration,
Table~\ref{tab:ablation} still shows two robust, large effects. Higher input
resolution rescues the tiny \emph{ifsc} field from near-zero IoU
(v1$\rightarrow$v2), and the $4{\times}2$ spatial pooling grid adds +0.15
mIoU (v2$\rightarrow$v3). The v2 column records a misstep worth reporting:
aggressive field weights ($4\times$ \emph{ifsc}, $3\times$ \emph{sign})
rescued the small fields but degraded the easy ones, dropping accuracy below
v1; v3 moderates the weights ($2\times$, $1.5\times$). These trends are
large enough to survive the seed noise quantified next; the absolute numbers
are not.

\paragraph{Does the synthetic data help? No.} Table~\ref{tab:controlled}
reports the controlled comparison, and it is a negative result. The
single-seed v4 of Table~\ref{tab:ablation} (0.704 mIoU, 98\% accuracy) is
the \emph{best} of three seeds; across all three the synthetic-augmented
model averages $0.648 \pm 0.041$ mIoU and $89 \pm 7\%$ accuracy, on mIoU
below the static prior. The compute-matched real-only control is decisive:
v3 trained for the same number of gradient steps (v3-long) reaches 0.660
mIoU and 93.3\% accuracy, matching or beating the synthetic-augmented model
on every aggregate metric while using no synthetic data at all. The apparent
v3$\rightarrow$v4 improvement in the single-seed ablation is therefore
explained by longer effective training and a favourable seed, not by the
synthetic images. The \emph{ifsc} field is the clearest illustration: its v4
accuracy swings 90/40/0\% across the three seeds (mean $43 \pm 37\%$),
whereas the real-only v3-long localises it at a stable 80\%, so the
``bank-conditional placement from synthetic data'' that a single run might
suggest does not hold up (per-bank IFSC placement is quantified in
Appendix~\ref{app:ifsc}).

\paragraph{Why might appearance augmentation not help here?} The cut-paste
pipeline adds appearance diversity (handwriting, ink, content) but, by the
same-coordinate paste of Section~\ref{sec:synthetic}, no \emph{layout}
diversity. The static baseline shows that layout is where the task's
difficulty and headroom lie; appearance is not the bottleneck, so augmenting
it does not move the metrics. This is a concrete caution for small-data
practitioners: cut-paste augmentation, effective for instance detection on
cluttered scenes~\cite{dwibedi2017cut}, need not transfer to fixed-layout
documents, and the productive direction is layout-varying synthesis
(paste-position jitter, additional bank templates) rather than more
same-layout composites.

\paragraph{What does help} Learning is not useless: with adequate training a
real-data model (v3-long) reaches 93\% accuracy at IoU $\geq 0.5$, well above
the prior's 80\%, even though it does not beat the prior on mean IoU. The
gains are concentrated in reducing gross localisation failures (the
accuracy metric) rather than in median box overlap. A single caveat
underlies all of this: the test set is 10 images and 60 boxes, so per-field
accuracies quantise in 10-point steps and even the seed-averaged aggregates
carry wide intervals; re-establishing these comparisons on the larger
canonical split (Section~\ref{sec:dataset}) is immediate future work.

\paragraph{Scope of this finding} Producing the strongest model was never
the aim of this work, and the negative result above is narrow: it says that
\emph{this} cut-paste data, with \emph{this} small regression baseline, on
\emph{this} 10-image test set, does not improve localisation. It does not
diminish the generation \emph{approach}, whose value is orthogonal to the
present experiment. The pipeline can synthesise an essentially unbounded
number of redistributable, fully-annotated cheque images from a handful of
real ones, which is exactly what is missing in this domain. Whether, and
how, such data improves model building, at larger scale, with stronger
architectures, with layout-varying synthesis, or as pre-training rather than
in-domain augmentation, is an open question that this study motivates but
does not settle, and that we leave to future work with the released
resources.

\begin{figure}[t]
  \centering
  \includegraphics[width=\linewidth]{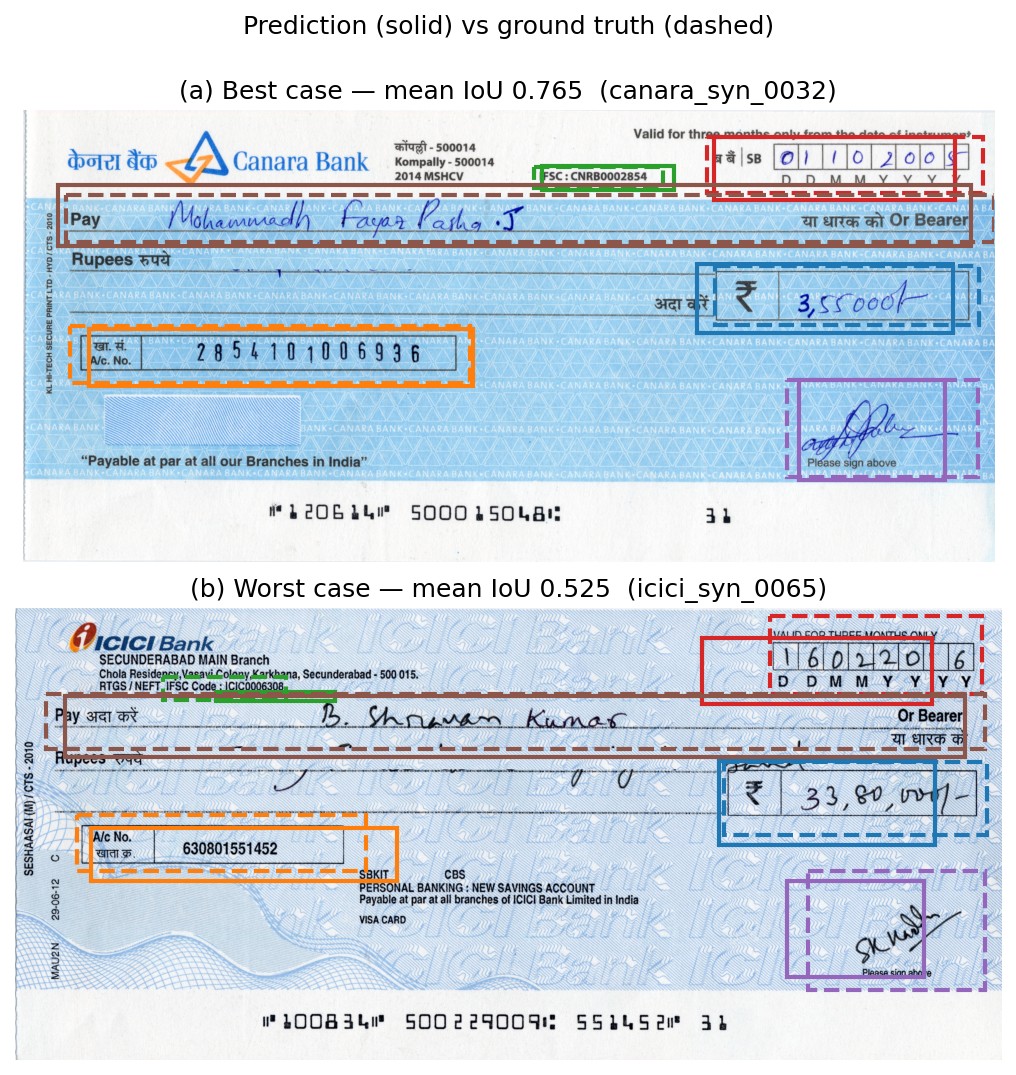}
  \caption{Qualitative v4 predictions (solid) vs.\ ground truth (dashed) on
  the synthetic test split, best (top) and worst (bottom) by mean IoU.}
  \label{fig:qualitative}
\end{figure}

Fig.~\ref{fig:qualitative} shows the best and worst predictions of the
(seed-42) v4 model on the synthetic test split, visualised there because the
real test images may not be reproduced here.

\section{Conclusion}
\label{sec:conclusion}

We released three resources that make the IDRBT cheque dataset usable for
modern machine learning: six-field annotations for all 112 images
(annotations-only, terms-respecting), a 295-image redistributable synthetic
dataset generated by coordinate-preserving cut-paste, and a ResNet-50
regression baseline together with the controlled experiment it enables. The
annotations and synthetic images are the lasting contributions; the baseline
serves to characterise the benchmark and to test the synthetic data
honestly. That test returned a negative result: a no-learning layout prior
already reaches 0.691 mIoU and 80\% accuracy, no learned variant beats it on
mIoU, and once seed variance and training compute are controlled, the
cut-paste synthetic data gives no measurable improvement over real data
alone. We attribute this to the method adding appearance but not layout
diversity, and we release the negative result as guidance rather than bury
it. Future work therefore targets layout-varying synthesis (paste-position
jitter, multiple canvas templates per bank, bank-conditioned models) and
re-evaluation on the larger canonical split. We hope these resources, and
the cautionary baseline, seed further work on cheque understanding, a domain
where public data has been nearly nonexistent.

\appendices

\section{Per-Bank IFSC Placement}
\label{app:ifsc}

Section~\ref{sec:results} attributes the static baseline's 0\% accuracy on
\emph{ifsc} to bank-dependent placement. Table~\ref{tab:ifsc} quantifies
this from the real annotations: within a bank the normalised IFSC centre-x
varies by at most $\pm 0.005$, but across banks it ranges from 0.172 (Axis)
to 0.590 (Canara). A single mean box sits at $\approx 0.227$, pulled off
the dominant Axis position (0.172) by the minority banks: with the field
only 0.125 wide, even that 0.055 offset caps the IoU near 0.39 on Axis
cheques, and the Canara and Syndicate placements (0.2--0.4 further away)
are missed entirely. The mean box therefore fails everywhere at the 0.5
threshold, which is precisely the structure a learned, bank-conditional
model can express and a static prior cannot.

\begin{table}[t]
  \caption{Normalised IFSC centre-$x$ by bank, from the real annotations.
  Bank labels come from the creators' registry (105 of 112 cheques); the
  Syndicate row uses the 6 registry-unmatched cheques identified through
  the synthetic provenance audit.}
  \label{tab:ifsc}
  \centering
  \small
  \begin{tabular}{lccc}
    \toprule
    Bank & $n$ & Centre-$x$ (mean) & Centre-$x$ ($\sigma$) \\
    \midrule
    Axis      & 87 & 0.172 & 0.005 \\
    Canara    & 10 & 0.590 & 0.003 \\
    ICICI     &  8 & 0.214 & 0.003 \\
    Syndicate &  6 & 0.395 & 0.003 \\
    \bottomrule
  \end{tabular}
\end{table}

\section{Synthetic Layout Fidelity}
\label{app:fidelity}

Section~\ref{sec:synthetic} claims the synthetic bounding-box statistics
closely match the real ones. Table~\ref{tab:fidelity} compares the
normalised centre and size statistics of the real annotations
(Table~\ref{tab:bbox_stats}) with those of the synthetic train split. The
coordinates agree to within rounding for every field except the \emph{ifsc}
centre-$x$, where the synthetic mixture mean shifts from 0.22 to 0.36 with
a large spread ($\pm 0.18$). This is expected, not an error: the
$10\times$ over-sampling of the small banks re-weights the bank mixture,
and as Table~\ref{tab:ifsc} shows, IFSC placement is the one strongly
bank-dependent coordinate. Within each bank the synthetic coordinates are
identical to the real ones by construction (same-position paste).

\begin{table}[t]
  \caption{Real vs.\ synthetic (train split) normalised box statistics:
  centre-$x$, centre-$y$, width, height (means; synthetic $\sigma$ in
  Section 5.4 of the data report shipped with the dataset).}
  \label{tab:fidelity}
  \centering
  \small
  \setlength{\tabcolsep}{4pt}
  \begin{tabular}{lcccccccc}
    \toprule
    & \multicolumn{2}{c}{Ctr-$x$} & \multicolumn{2}{c}{Ctr-$y$}
    & \multicolumn{2}{c}{W} & \multicolumn{2}{c}{H} \\
    Field & real & syn & real & syn & real & syn & real & syn \\
    \midrule
    date   & 0.854 & 0.86 & 0.125 & 0.12 & 0.264 & 0.26 & 0.136 & 0.14 \\
    amount & 0.846 & 0.85 & 0.424 & 0.42 & 0.269 & 0.27 & 0.135 & 0.14 \\
    ifsc   & 0.224 & 0.36 & 0.157 & 0.16 & 0.125 & 0.13 & 0.047 & 0.05 \\
    acno   & 0.211 & 0.24 & 0.529 & 0.53 & 0.317 & 0.36 & 0.107 & 0.12 \\
    sign   & 0.904 & 0.89 & 0.721 & 0.73 & 0.140 & 0.15 & 0.221 & 0.21 \\
    name   & 0.510 & 0.51 & 0.251 & 0.25 & 0.960 & 0.95 & 0.109 & 0.11 \\
    \bottomrule
  \end{tabular}
\end{table}

\section{Reproducibility Details}
\label{app:repro}

\paragraph{Software and hardware} PyTorch 2.7.1 (CUDA 12.6 build),
torchvision 0.22.1, on a single NVIDIA GeForce GTX 1060 (6\,GB). A full
150-epoch v4 run takes roughly 4.5--5 hours at batch size 4.

\paragraph{Determinism} The real train/val/test split is fixed by a seeded
\texttt{random\_split} (seed 42) over the 105 valid HDF5 rows; synthetic
generation and its splits use seed 42. Training scripts accept a separate
\texttt{--model-seed} that re-seeds initialisation and augmentation
\emph{after} the split is created, so replicate runs share the identical
evaluation data.

\paragraph{Exact commands} The v4 model of Table~\ref{tab:ablation} is
produced by

{\small
\begin{verbatim}
python train_resnet.py \
  --save-dir models/resnet_cheque_v4_synth \
  --synthetic-train <synthetic train parquet> \
  --synthetic-exclude synthetic_exclude_ids.txt
\end{verbatim}
}

\noindent the cross-annotator agreement of Section~\ref{sec:agreement} by
\texttt{python paper/annotation\_agreement.py},
the seed replicates by adding \texttt{--model-seed 43} (and 44),
the compute-matched control by \texttt{python train\_resnet.py --epochs 495}
(real data only), the static baseline by
\texttt{python paper/static\_baseline.py}, and the leakage audit by
\texttt{python synthetic\_provenance.py}. Each training run writes
\texttt{metrics.json}, \texttt{config.json}, and training curves next to its
checkpoint; Table~\ref{tab:controlled} aggregates these.

\paragraph{Checkpoint provenance} The released v4 checkpoint is the seed-42
run, which is the best of the three seeds in Table~\ref{tab:controlled}; we
release it as a usable model while reporting the seed-averaged result as the
scientific finding. It corresponds to epoch 141 (best validation mIoU
0.6513); the run was interrupted at epoch 113 by a machine fault and resumed
from its epoch-105 best checkpoint with \texttt{--resume --start-epoch 106},
which restarts the optimiser at learning rate $10^{-5}$ with cosine decay
over the remaining epochs, so the post-resume schedule differs slightly from
an uninterrupted run. We disclose this because the checkpoint is public and
the training log is part of the repository history.

\section*{Ethics and Generative AI Disclosure}
This work revives a personal project that had been abandoned for roughly six
years. The original codebase, including the annotation effort and the early
processing pipeline, was written before generative AI tooling existed. In
2026 the author resumed the project and discloses the use of generative AI
assistants, namely GitHub Copilot and Claude Code, to generate and refactor
code, write test cases, and augment documentation in preparing the resources
and this paper for publication. All annotations were created and verified
manually by the author (Section~\ref{sec:dataset}); experimental results were
produced by the released code and reviewed by the author, who takes full
responsibility for the content of this paper.

\section*{Acknowledgements}
The IDRBT Cheque Image Dataset~\cite{idrbt2020} is provided by the Institute
for Development and Research in Banking Technology, Hyderabad.

\bibliographystyle{IEEEtran}
\bibliography{references}

\end{document}